\documentclass[conference]{IEEEtran}
\pdfoutput=1
\IEEEoverridecommandlockouts
\usepackage{cite}
\usepackage{amsmath,amssymb,amsfonts}
\usepackage{algorithmic}
\usepackage{graphicx}
\usepackage{textcomp}
\usepackage{xcolor}
\def\BibTeX{{\rm B\kern-.05em{\sc i\kern-.025em b}\kern-.08em
    T\kern-.1667em\lower.7ex\hbox{E}\kern-.125emX}}
\begin{document}

\title{Sci-CoT: Leveraging Large Language Models for Enhanced Knowledge Distillation in Small Models for Scientific QA\\
}

\author{\IEEEauthorblockN{Yuhan Ma}
\IEEEauthorblockA{\textit{South China Normal University} \\
Guangdong 510632, China \\
2022024953@m.scnu.edu.cn}
~\\
\and
\IEEEauthorblockN{Chenyou Fan*}
\IEEEauthorblockA{\textit{South China Normal University} \\
Guangdong 510632, China \\
fanchenyou@scnu.edu.cn}

~\\
\and
\IEEEauthorblockN{Haiqi Jiang}
\IEEEauthorblockA{\textit{South China Normal University} \\
Guangdong 510632, China \\
20214001099@m.scnu.edu.cn}
}

\maketitle{}

\begin{abstract}

Large Language Models (LLMs) have shown outstanding performance across wide range of downstream tasks. This competency is attributed to their substantial parameter size and pre-training on extensive corpus. Moreover, LLMs have exhibited enhanced reasoning capabilities in tackling complex reasoning tasks, owing to the utilization of a method named ``Chain-of-Thought (CoT) prompting''. This method is designed to generate intermediate reasoning steps that guide the inference of the final answer. However, it is essential to highlight that these advanced reasoning abilities appear to emerge in models with a minimum of 10 billion parameters, thereby limiting its efficacy in situations where computational resources are constrained.
In this paper, we investigate the possibility of transferring the reasoning capabilities of LLMs to smaller models via knowledge distillation. Specifically, we propose Sci-CoT, a two-stage framework that separates the processes of generating rationales and inferring answers. This method enables a more efficient use of rationales during the answer inference stage, leading to improved performance on scientific question-answering tasks. 
Utilizing Sci-CoT, our 80-million parameter model is able to exceed the performance of BLOOM-176B in the ARC-Easy dataset under the few shot setting.
\end{abstract}


\section{Introduction}

Recently, CoT prompting methods for LLMs have made it possible for them to decompose complex reasoning tasks into a series of intermediate steps~\cite{b1}~\cite{b19}. This advancement has significantly improved the reasoning capabilities of LLMs, enabling them to excel in a variety of downstream tasks and even achieve state-of-the-art results in specific tasks. Wei~et al.~\cite{b1} discovered that these CoT prompting methods are effective only for models with at least 10 billion of parameters, such as GPT-3 (175B)~\cite{b2} and PaLM (540B)~\cite{b3}. Conversely, for models with less than 10 billion parameters, these prompting methods not only fails to enhance performance but can potentially impair it. According to Wei et al.~\cite{b1}, small models, specifically those with fewer than 10 billion parameters, exhibit a deficiency in reasoning capabilities. This shortfall is identified as the primary reason for their poor performance when employing CoT prompting methods. Therefore, our research focused on employing a knowledge distillation method to enable smaller models to learn the reasoning capabilities of the LLMs, ultimately to improve the performance of small models.

In this study, we propose \emph{Sci-CoT}, a novel framework aimed at transferring the reasoning capabilities of LLMs to smaller student models. This process serves to improve their performance specifically in the realm of scientific question-answering tasks. While similar attempts have been made in recent studies, such as those by Magister~et al.~\cite{b3} and Ho~et al.~\cite{b4}, we aim to address several gaps in our paper. Specifically, previous work tends to (\romannumeral1) focus predominantly on mathematical problems and (\romannumeral2)  implement a one-stage method that uses a question as input and using LLMs to generate explanations (rationales) for each question, coupled with the correct answers, serve as labels for smaller student models to learn. However, our experimental results indicate that this one-stage method does not effectively improve the performance of our small student model in the context of scientific question-answering tasks. Furthermore, previous studies (\romannumeral3) commonly utilize small models that are still fairly computationally intensive. With parameter size ranging from 1 to 11 billion, these models pose a significant challenge for deployment in environments where computational resources are limited.

\begin{figure}[htbp] 
\centering 
\includegraphics[width=0.5\textwidth]{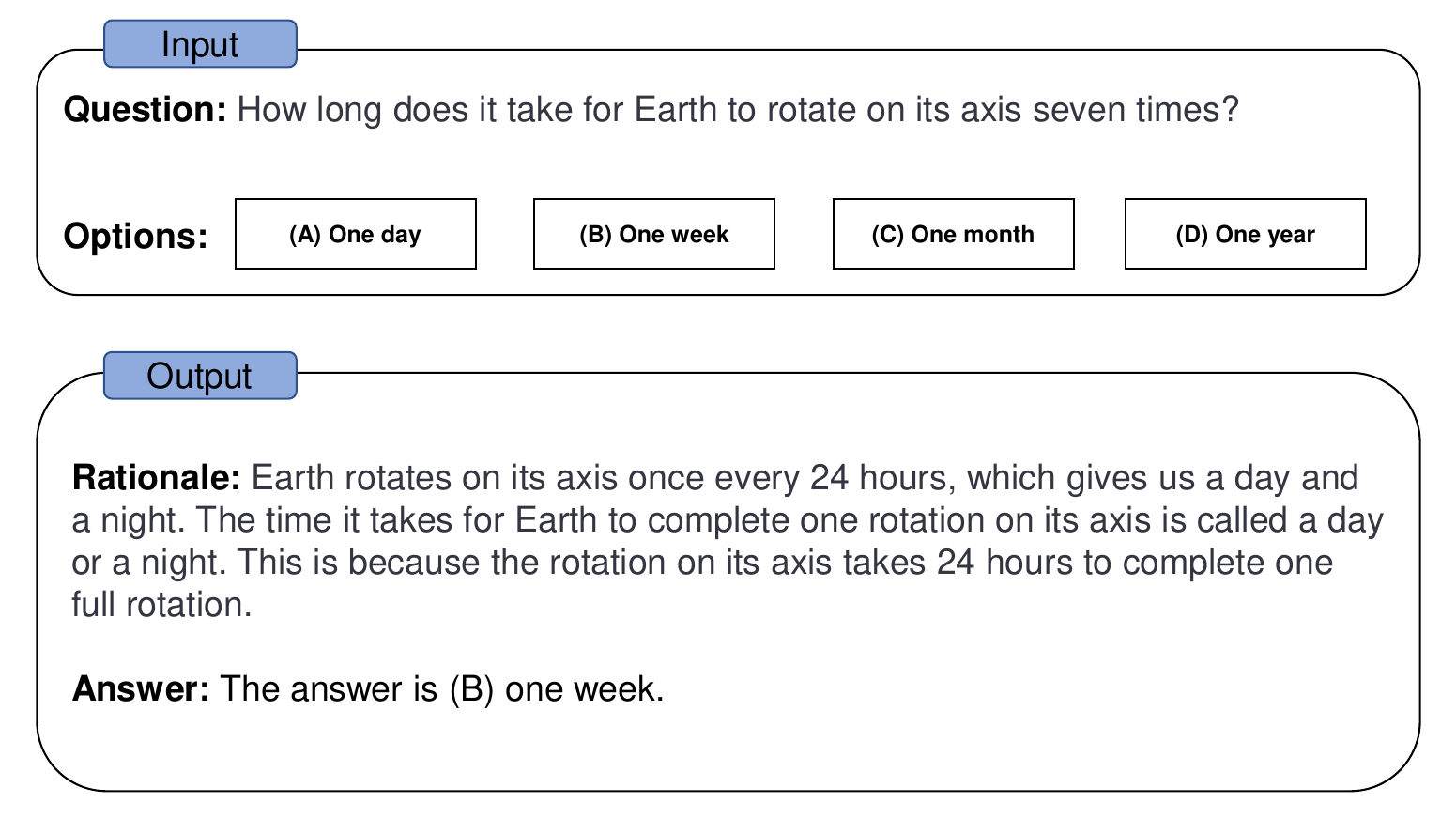}
\caption{Example of the Sci-CoT task} 
\label{intro.pdf} 
\end{figure}

In mathematical problems, the evaluation of the correctness of a rationale is relatively straightforward. This typically involves a comparison of the output answers with the established ground truth. If the outputs coincide with the ground truth, the corresponding rationale is deemed correct. But when it comes to science question-answering problems, figuring out whether rationales are accurate isn't that simple. As a solution, we provide LLMs with the correct answer accompanied by the prompt ``Let's think step by step''~\cite{b5}, as a strategy to encourage the generation of accurate rationales for each question. 
The student model we utilized in our study, Flan-T5-small, contains 80 million parameters, a number that is considerably smaller than those in previous studies. This reduction in parameters limits the model's reasoning capability. To fully leverage this student model, we introduce Sci-CoT, a method involving two distinct stages. An example of a Sci-CoT task is illustrated in Figure \ref{intro.pdf}. Upon receiving a question, we initially generate a rationale for it. This question and its corresponding rationale are then concatenated and used as input to perform answer inference.
With Sci-CoT, our student model can achieve better performance on scientific question-answering tasks. 

We summarize our contribution as follows:
\begin{enumerate}
    \item Prior research has primarily focused on mathematical problems. However, our study uniquely targets scientific question-answering tasks, thus establishing a paradigm distinct from those seen in previous studies.
    \item We propose a novel two-stage framework designed to enhance the reasoning abilities of the student model. Utilizing this framework, our student model demonstrates superior performance on the Arc-Easy and Arc-Challenge datasets. Remarkably, our student model outperforms BLOOM-176B on the ARC-Easy dataset and achieves comparable accuracy to the OPT-175B on the Arc-Challenge dataset.
\end{enumerate}

\section{Related Work}

\subsection{CoT Reasoning with LLMs}
Recently, the CoT prompting methods have gained considerable attention for its capacity to facilitate multi-step reasoning in LLMs~\cite{b1}. Specifically, CoT prompting approaches empower LLMs to generate sequential reasoning chains, serving as an effective problem-solving tool. Researches have demonstrated that LLMs can employ CoT reasoning through two main paradigms: Zero-Shot-CoT~\cite{b5} and Few-Shot-CoT~\cite{b1}~\cite{b18}. In the case of Zero-Shot-CoT, studies have indicated that LLMs exhibit competent zero-shot reasoning capabilities when prompted with specific phrases like ``Let's think step by step'' following a question. Conversely, Few-Shot-CoT applies a limited number of step-by-step reasoning demonstrations as conditioning factors for inference. Each demonstration comprises a question paired with a reasoning chain that leads to the final answer. These demonstrations are typically acquired either via manual composition or automated generation.

\subsection{Knowledge Distillation}
Hinton et al.~\cite{b6}  were the pioneers in introducing Knowledge Distillation (KD), a methodology that involves the training of streamlined models, which are derived from larger and more intricate models. The primary objective of KD is to minimize model size and latency, without compromising accuracy or the capacity for generalization. This method is particularly beneficial for deployment on devices with restricted resources.
Among various KD techniques, one of the most prevalent involves instructing the student model with the aim of mirroring the teacher model's representation, which includes elements such as logits and the output probability of the succeeding word. Distill-BERT~\cite{b7} utilized KD during the pre-training phase, BERT-PKD~\cite{b8} applied KD  during the fine-tuning stage, and Tiny-BERT~\cite{b9} employed KD in both pre-training and fine-tuning stages.

\subsection{Eliciting CoT Reasoning by Fine-Tuning Models}
Recent researches have seen a surge of interest in eliciting CoT reasoning through the fine-tuning of language models. Lu et al.~\cite{b10} fine-tuned a T5 model on ScienceQA dataset, which includes intermediate reasoning step annotations. However, the model's performance declined when leveraging reasoning steps for answer inference. 
Similarly, Ho et al.~\cite{b4} and Magister et al.~\cite{b3} employed knowledge distillation techniques to fine-tune a student model based on the CoT outputs generated by a more extensive teacher model. Their proposed methodologies demonstrated significant performance enhancements in arithmetic, and symbolic reasoning tasks. 

\section{Method}
\begin{figure}[htbp] 
\centering 
\includegraphics[width=0.5\textwidth]{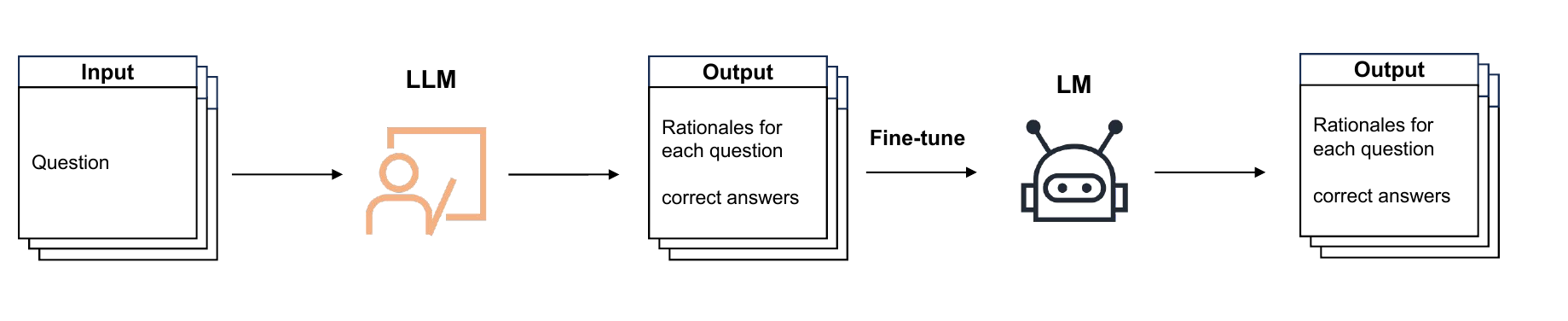}
\caption{\textbf{Overview of the one-stage method employed by prior related research.} Initially, this method employs a LLM as the teacher model, which takes the question as input and is tasked with generating rationales and correct answers concurrently for each question. Subsequently, a smaller Language Model (LM) is utilized as the student model, which is fine-tuned using the question as input and the outputs from the LLM as labels. Finally, this LM is deployed to generate both the rationale and answer inference simultaneously.}
\label{Overview of one-stage method utilized by prior related researches} 
\end{figure}

\begin{figure}[htbp] 
\centering 
\includegraphics[width=0.5\textwidth]{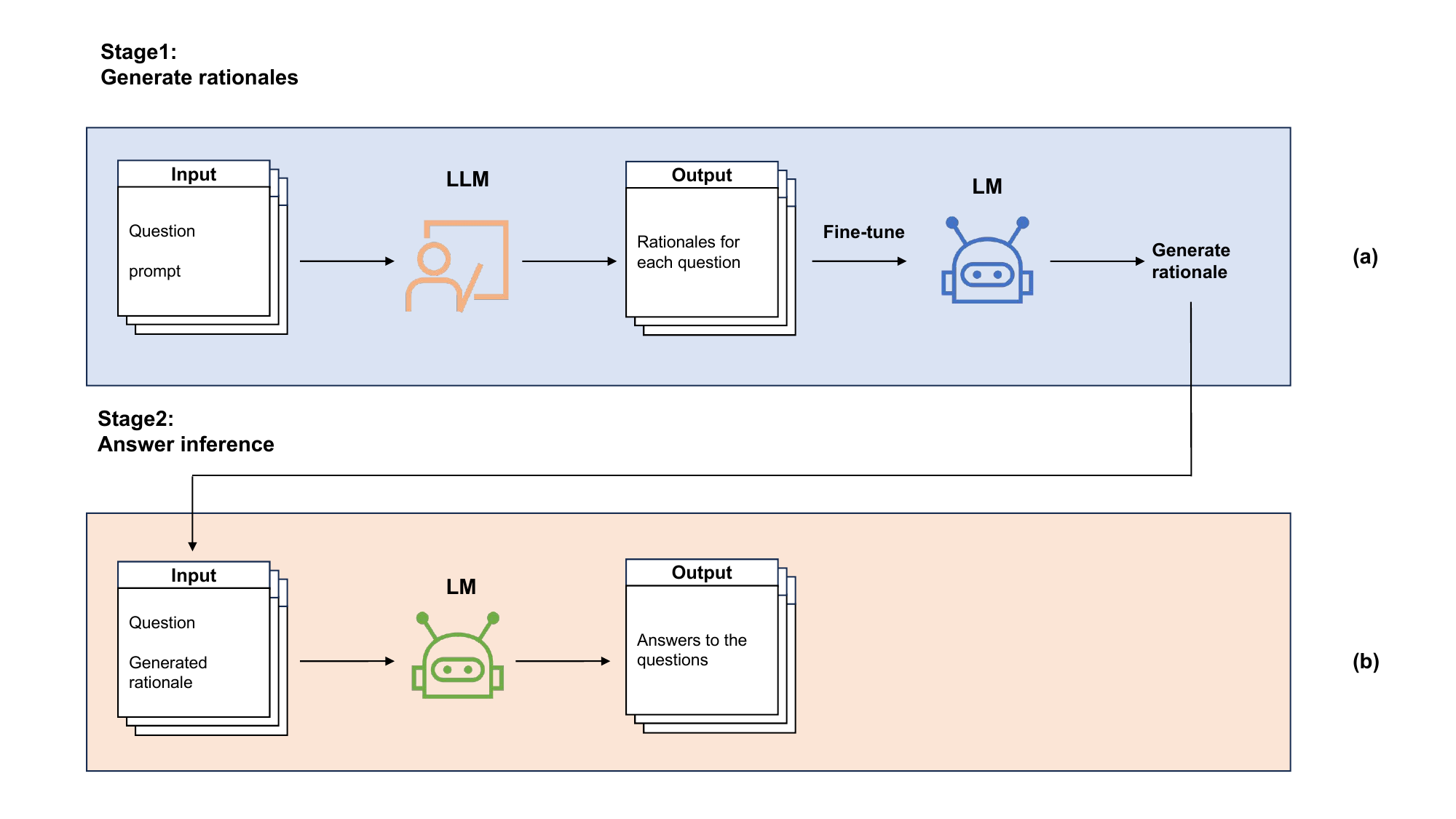}
\caption{\textbf{Overview of Sci-CoT framework.} Our proposed framework comprises two distinct stages: (\romannumeral1) rationale generation and (\romannumeral2) answer inference, distinguishing it from the prior one-stage method. Our results confirm that our framework significantly outperforms the one-stage method.}
\label{method.pdf} 
\end{figure}

In this section, we present an overview of the one-stage method previously employed in the related research, as is illustrated in Figure \ref{Overview of one-stage method utilized by prior related researches}. Furthermore, we introduce our novel Sci-CoT framework, distinct for its two-stage structure, as demonstrated in Figure \ref{method.pdf}. Additionally, we provide a concise overview of the teacher and student models deployed in our study.

\subsection{Sci-CoT}

We illustrate our Sci-CoT framework through two distinct stages, as depicted in Figure \ref{method.pdf}.

\textbf{Stage 1. Generate rationale}
In the initial stage of our methodology, we employ a large language model (LLM) to act as a teacher model, generating detailed explanations that serve as rationales for each posed question. To maintain the integrity and accuracy of these machine-generated rationales, we incorporate two uniquely constructed prompts. The first prompt embodies the correct answer to the respective question, thereby providing a reliable reference point. Meanwhile, the second prompt, ``Let's think step by step''~\cite{b5}, encourages analytical reasoning. 

In addition to these, we deploy a third, specialized prompt that directs the LLM to generate unique rationales for each distinct query.  Furthermore, to ensure that the generated rationales solely encompass explanations to the questions, we undertake a rigorous process of manual data inspection and cleaning.

Subsequently, we use the question as input along with the generated rationales as labels, fine-tune a smaller student model. Ultimately,, this small student model is then employed to generate appropriate rationales for each question. A visual representation of our method is provided in Figure \ref{method.pdf} (a). The mathematical representation of the loss function, is displayed in Equation \ref{eq:loss_function1}. In this equation, the symbol $\ell$ represents the cross-entropy loss between the predicted tokens and their corresponding target tokens. Here, $N$ indicates the length of the target sequences, whereas $i$ denotes the $i_{th}$ position in both the predicted and target tokens. The decision to utilize cross-entropy as our loss function was inspired by the work of Hsieh et al.~\cite{b11}.

\begin{equation}
\mathcal{L}_{rationale} = \frac{1}{N}\sum_{i=1}^{N}\ell(y_i, \hat{y_i})
\label{eq:loss_function1}
\end{equation}

\textbf{Stage 2. Answer inference}
During the answer inference stage, we take the question and the rationale produced by the student model in the first stage as inputs, using the correct answers as labels, to fine-tune an another smaller model. This small language model  is subsequently employed to carry out the answer inference task, as illustrated in Figure \ref{method.pdf} (b).

Significantly, this model possesses the same architecture and parameter size as the student model employed in the first stage. While these two small models share the same architecture, their applications differ. The model in the first stage is designed to generate rationales for each question. Conversely, in this stage, the model is purposed for executing the answer inference task.

The loss function applicable to this phase is articulated in Equation \ref{eq:loss_function2}. In this equation, the symbol $\ell$ stands for the cross-entropy loss, determined by comparing the model's predicted outputs with their corresponding target outputs. The variable $N$ represents the length of the target sequences, and $i$ refers to the $i_{th}$ position within both the predicted and target token sequences.

\begin{equation}
\mathcal{L}_{answer} = \frac{1}{N}\sum_{i=1}^{N}\ell(y_i, \hat{y_i})
\label{eq:loss_function2}
\end{equation}
\subsection{LLM Teacher model}

In our research, we have utilized GPT-3.5-turbo~\cite{b12} as the teacher model for our study. This model is provided through the OpenAI's API service.  This choice is based on the model's natural ability to create clear and relevant responses, a capability gained from its training using unsupervised learning methods on a wide range of different texts.
\subsection{Student model}
For this study, We have selected Flan-t5-small~\cite{b13} as our student model. Flan-T5-small is a compact variant of T5 (Text-to-Text Transfer Transformer)~\cite{b14} model, designed to provide robust natural language processing capabilities while maintaining a smaller size and lower computational demands. 
The architecture of the Flan-T5-small is  built upon the Transformer model, which takes advantage of self-attention mechanisms to efficiently process and generate text. With fewer layers and parameters than its larger counterparts, Flan-T5-small is ideal for resource-limited environments or applications that require a balance between performance and resource usage. 
Despite its compact size, Flan-T5-small retains the ability to perform diverse tasks, including text summarization, question-answering, and translation, demonstrating its versatility and power. 

\section{Experiment}
\label{sec:exp}
We assess the performance of our proposed Sci-CoT framework using the benchmark datasets ARC-Easy and ARC-Challenge.

Our evaluation involves a comparative analysis between our proposed framework and the conventional fine-tuning approach. Furthermore, we contrast our methodology with a previously used one-stage method, which  involves using the question as input and generating rationales along with the correct answers as output. The results of our study indicate that the Sci-CoT method exhibits markedly improved performance in comparison to these other approaches.

\subsection{Datasets}
\label{sec:dataset}
\textbf{ARC-Easy}~\cite{b15}: This dataset is the subset of the AI2 Reasoning Challenge (ARC). This dataset comprises 5,197 natural science questions. These multiple-choice questions, originating from elementary and middle school science exams, span a wide array of topics including nature and fundamental sciences.

\textbf{ARC-Challenge}~\cite{b15}: This dataset is also subset of the AI2 Reasoning Challenge (ARC) and contains 2590 natural science questions. encompasses 2,590 natural science questions. While the question structure aligns with that of ARC-Easy, this particular subset consists of more complex, intellectually challenging questions.

To simplify the expression, we employ the abbreviations ``ARC-E" and ``ARC-C" to respectively denote the ARC-Easy dataset and the ARC-Challenge dataset.

\subsection{Results of Sci-CoT}
\label{sec:Results of Sci-Fine-tune-CoT}
As a gentle reminder, in the subsequent discussions, Sci-CoT refers to the process of leveraging large language models to distill reasoning abilities into a significantly smaller student model using a two-stage framework, thereby enhancing the performance of the smaller model. 

Moreover, for comparative purposes, we  set the conventional fine-tune method as our baseline, which involves using questions as input and their corresponding correct answers as labels. This technique involves the utilization of questions as input data and the corresponding correct answers as labels.  Further, we contrast our method with results obtained from various large language models. These comparative results are presented in Table \ref{tab:Results of Sci-CoT}. 

\begin{table}[htbp]
\caption{Results of Sci-CoT}
\begin{center}
\begin{tabular}{|c|c|c|c|}
\hline
\textbf{Model}&\textbf{Model Size}&\textbf{ARC-E}&\textbf{ARC-C} \\
\cline{1-4} 
OPT (50\% Sparsity)~\cite{b16} & 175B & 28.03\% & 25.60\% \\
\hline
OPT (few-shot k=5)~\cite{b17} & 175B & 37.40\% & 31.10\% \\
\hline
BLOOM (few-shot k=5)~\cite{b17} & 176B & 40.70\% & \textbf{32.90}\% \\
\hline
Fine-tune (Baseline) & 80M & 38.04\% & 25.25\% \\
\hline
Sci-CoT (ours) & 80M & \textbf{43.73\%} & 31.05\% \\
\hline
\end{tabular}
\label{tab:Results of Sci-CoT}
\end{center}
\end{table}

\textbf{Small models can outperform large language models.} The results presented in Table \ref{tab:Results of Sci-CoT} demonstrate the effectiveness of our proposed framework, particularly in the ARC-Easy dataset. Although we use a small model with just 80M parameters, our approach achieve superior performance compared to the larger OPT and BLOOM models. Interestingly, even on the ARC-Challenge dataset, which questions are more challenging, our small model manages to achieve results comparable to those of OPT and BLOOM. These findings underscore the potential for smaller models to compete with, and even outperform their larger counterparts when appropriately configured and optimized.

\textbf{Sci-CoT exceeds conventional fine-tune method.} Upon comparing our approach to the conventional fine-tuning method on both the ARC-Easy and ARC-Challenge datasets, we observe notable improvements achieved by Sci-CoT. Specifically, on the ARC-Easy dataset, the implementation of our proposed framework led to an improvement of 5.69\% (38.04\%$\rightarrow$43.73\%). Likewise, the ARC-Challenge dataset also showed a 5.8\% enhancement in performance (25.25\%$\rightarrow$31.05\%). These results exhibit the effectiveness of our Sci-CoT framework.

\subsection{Sci-CoT vs One-stage method}
\label{sec:Sci-CoT vs One-stage method}

\begin{figure}[htbp] 
\centering 
\includegraphics[width=0.5\textwidth]{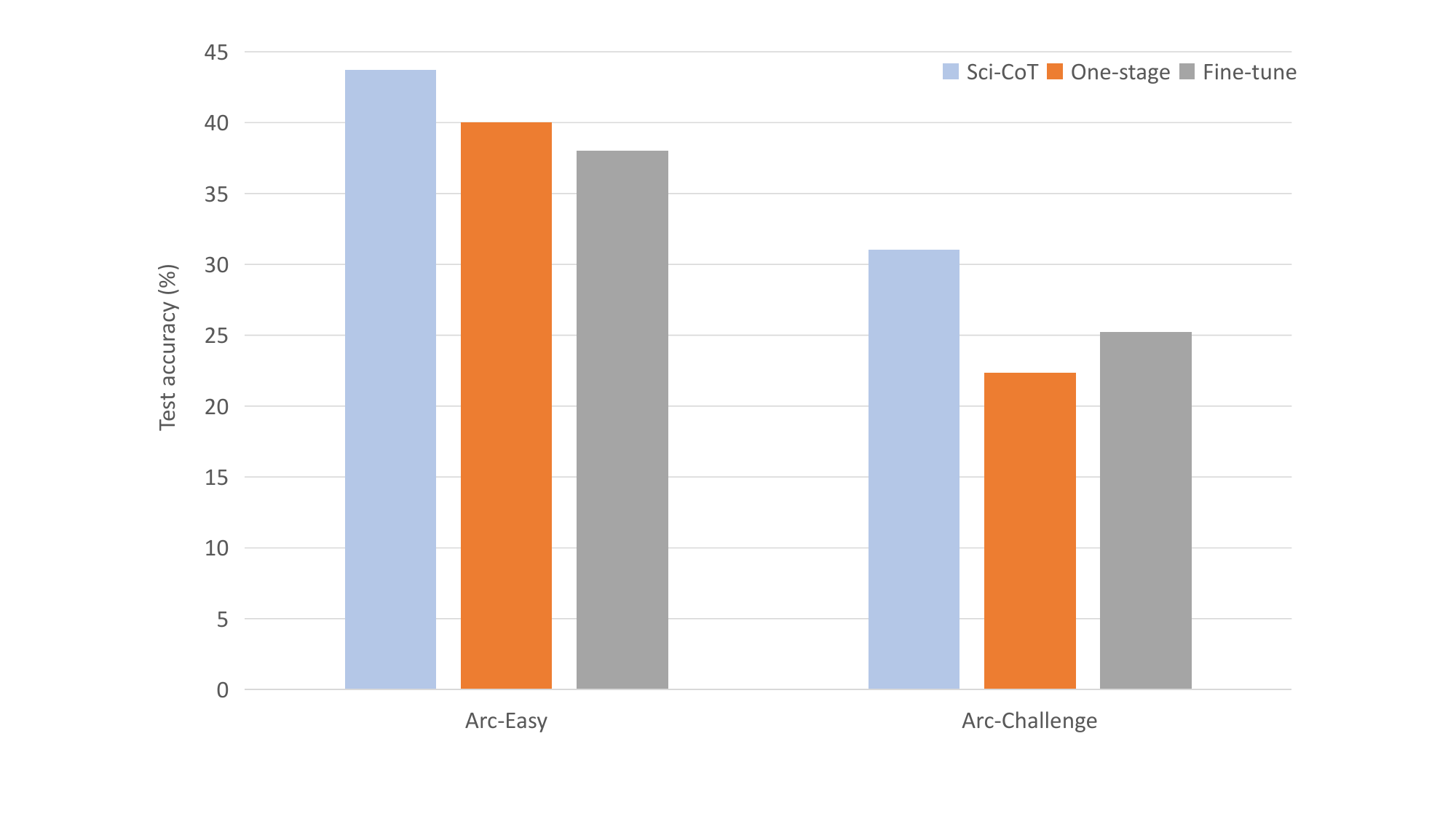} 
\caption{\textbf{Ablation study on one-stage method}} 
\label{Ablation study on one-stage method} 
\end{figure}

Within the framework of Sci-CoT, we propose a two-stage framework, which involves utilizing two models with identical architecture but for distinct purposes. The first model generates CoT for each question, while the second model performs answer inference. This two-stage algorithm is then contrasted with a one-stage method that employs a single model to both generate CoT and infer subsequent answers. The results are shown in Table \ref{tab:Ablation study on one-stage method} and Figure \ref{Ablation study on one-stage method}.

\begin{table}[htbp]
\caption{Ablation study on one-stage method}
\begin{center}
\begin{tabular}{|c|c|c|}
\hline
\textbf{Method}&\textbf{ARC-E}&\textbf{ARC-C} \\
\cline{1-3} 
\hline
Fine-tune & 38.04\% & 25.25\% \\
\hline
One-stage  & 40.02\% & 22.35\% \\
\hline
Sci-CoT (ours)  & \textbf{43.73\%} & \textbf{31.05\%} \\
\hline
\end{tabular}
\label{tab:Ablation study on one-stage method}
\end{center}
\end{table}

From Table \ref{tab:Ablation study on one-stage method} and Figure \ref{Ablation study on one-stage method}, it is clear that Sci-CoT does better than the one-stage method for both ARC-Easy and ARC-Challenge datasets. Notably, while the one-stage method improves performance compared to the conventional fine-tuning approach on the ARC-Easy dataset, it fails to surpass the results achieved by Sci-CoT. In a more challenging setting, on the ARC-Challenge dataset, the one-stage method performs even worse than the fine-tune method. This inferior performance might be attributed to the limited parameter size of our chosen model, Flan-T5-small, which likely restricts its ability to concurrently generate rationales and infer answers.

\subsection{Ablation study on training data size}

\begin{figure}[htbp] 
\centering
\includegraphics[width=0.5\textwidth]{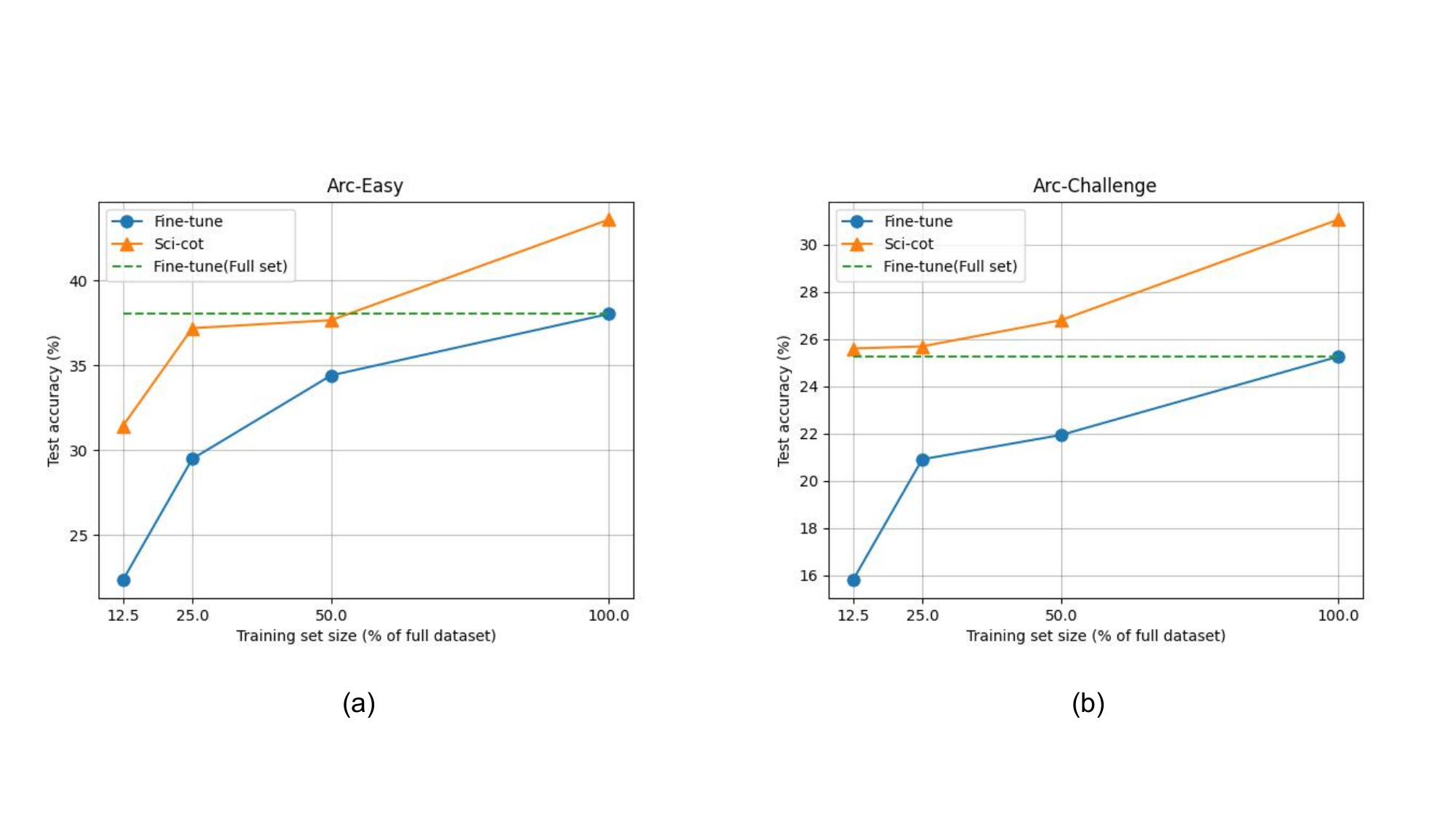} 
\caption{\textbf{Ablation study on training data size}} 
\label{Ablation study on training data size} 
\end{figure}

As we see improvements with our proposed Sci-CoT, we conducted an analysis to understand the trade-off between this improved performance and the size of the dataset used. For this, we performed experiments with different dataset sizes and compared the performance of Sci-CoT against the conventional fine-tune method. The results can be seen in Figure \ref{Ablation study on training data size}. Comprehensive data are articulated in Table \ref{tab:Ablation study on training data size (ARC-Easy)} and Table \ref{tab:Ablation study on training data size (ARC-Challenge)}.

\begin{table}[htbp]
\caption{Ablation study on training data size (ARC-E)}
\begin{center}
\begin{tabular}{|c|c|c|}
\hline
\textbf{Percentage of ARC-E used to train}&\textbf{Fine-tune}&\textbf{Sci-CoT} \\
\cline{1-3} 
\hline
12.5\% (281 examples) & 22.34\% & 31.43\% \\
\hline
25\% (562 examples)  & 29.50\% & 37.20\% \\
\hline
50\% (1125 examples) & 34.42\% & 37.66\%\\
\hline
100\% (2251 examples) & 38.04\% & 43.73\%\\
\hline
\end{tabular}
\label{tab:Ablation study on training data size (ARC-Easy)}
\end{center}
\end{table}

\begin{table}[htbp]
\caption{Ablation study on training data size (ARC-C)}
\begin{center}
\begin{tabular}{|c|c|c|}
\hline
\textbf{Percentage of ARC-C used to train}&\textbf{Fine-tune}&\textbf{Sci-CoT} \\
\cline{1-3} 
\hline
12.5\% (139 examples) & 15.78\% & 25.59\% \\
\hline
25\% (279 examples)  & 20.90\% & 25.68\% \\
\hline
50\% (559 examples) & 21.92\% & 26.79\%\\
\hline
100\% (1119 examples) & 25.25\% & 31.05\%\\
\hline
\end{tabular}
\label{tab:Ablation study on training data size (ARC-Challenge)}
\end{center}
\end{table}

Figure \ref{Ablation study on training data size} (a) indicates that Sci-CoT requires only about 50\% of the training data to reach the performance achieved by the fine-tune method when utilizing 100\% of the training data. Figure \ref{Ablation study on training data size} (b) further highlights that a satisfactory performance can be obtained using merely 25\% of the training data. This observation could possibly be attributed to the insufficient training data in the ARC-Challenge, which consists of a total of only 1,119 training samples.

\section{Conclusion and Future Work}
In this paper, we explored the method of transferring the reasoning capabilities of LLMs to a small model, aiming to enhance the performance of the small model in scientific question-answering tasks. Experimental results on the ARC-Easy and ARC-Challenge datasets revealed that this method can effectively learn from the reasoning capabilities of the LLMs, thus significantly augmenting the performance of the small model.

Future research may cover a wider range of reasoning tasks, including symbolic reasoning and logical reasoning.  A potential research direction is to explore better utilizing and unifying distinct existing LLMs to construct a more powerful and general open-source model through knowledge distillation and other methodology.

\end{document}